\documentclass[letterpaper, 10pt, conference]{ieeeconf}      

\IEEEoverridecommandlockouts                              

\usepackage{amsmath} 
\usepackage{amssymb}  
\usepackage{todonotes}
\let\labelindent\relax 
\usepackage{enumitem}
\usepackage{mathtools}
\usepackage{rotating}
\usepackage{multirow}
\usepackage{siunitx}
\usepackage{caption}
\usepackage{subcaption}
\usepackage{float}
\usepackage{hyperref}
\usepackage{pifont}
\usepackage{cite}

\usepackage{booktabs} 

\usepackage{algorithmic} 
\usepackage{algorithm2e}
\graphicspath{{graphics/}}

\usepackage{tabto} 

\usetikzlibrary{matrix, positioning, calc}
\tikzset{
  quadratic/.style={
    to path={
      (\tikztostart) .. controls
      ($#1!1/3!(\tikztostart)$) and ($#1!1/3!(\tikztotarget)$)
      .. (\tikztotarget)
    }
  }
}

\hypersetup{
    colorlinks=true,
    linkcolor=black,
    filecolor=magenta,      
    urlcolor=cyan,
    pdftitle={Object Safety Metric},
    pdfpagemode=FullScreen,
    }

\title{\LARGE \bf
A Comprehensive Safety Metric to \\ Evaluate Perception in Autonomous Systems
}

\author{Georg Volk$^{1}$, J\"org Gamerdinger$^{1}$, Alexander von Bernuth$^{1}$ and Oliver Bringmann$^{1}$
\thanks{$^{1}$University of T\"ubingen, Faculty of Science, Department of Computer Science, Embedded Systems Group \tt\small {\{volkg, gamerdjo, bernuth, bringmann\} @informatik.uni-tuebingen.de}}%
}

\begin{document}

\maketitle
\thispagestyle{empty}
\pagestyle{empty}

\begin{abstract}

Complete perception of the environment and its correct interpretation is crucial for autonomous vehicles. Object perception is the main component of automotive surround sensing. Various metrics already exist for the evaluation of object perception. However, objects can be of different importance depending on their velocity, orientation, distance, size, or the potential damage that could be caused by a collision due to a missed detection. Thus, these additional parameters have to be considered for safety evaluation. We propose a new safety metric that incorporates all these parameters and returns a single easily interpretable safety assessment score for object perception. This new metric is evaluated with both real world and virtual data sets and compared to state of the art metrics.

\end{abstract}

\section{INTRODUCTION}

The development of autonomous vehicles is being pushed forward by many manufacturers and research institutes. It is expected to bring various benefits, such as reducing the number of accidents and traffic jams. However, before marketability is reached, the safety of self-driving cars must be guaranteed. In order to be able to carry out and plan safe actions, an autonomous vehicle must be able to perceive its environment completely and correctly. Therefore, new methods to verify the safety of a perception system are needed.
All sensors must work correctly and have to be resilient against environmental influences for the sensed environment to be trusted. Furthermore, the system must process these data taking the environment, the traffic situation, and the properties of all road users into account.
Unfortunately, those systems do not always work as expected to this date.
This is shown by fatal accidents caused by self-driving cars, where the
``self-driving system software classified the pedestrian as an
unknown object, as a vehicle, and then as a bicycle with varying expectations of future travel path''~\cite{uber-accident}.

Accidents like this could have been avoided if the conducted tests had 
covered possible environmental conditions more completely. Additionally, the
metrics used to evaluate such systems should have paid
more attention to progression in time, possible vulnerability of third party
road users, or consistency of the perceived objects and their classifications.
Existing metrics do not provide any information about the safety of an
entire perception system including object criticality and environmental conditions.

In this work, we propose a metric that takes all these parameters into
account and provides a simple yet powerful measure of safety for object detection and
tracking algorithms.
In Section~\ref{sec:related}, we examine important existing but functionally
lacking metrics.
Then, we present our improved safety metric in Section~\ref{sec:metric}.
Finally, in Section~\ref{sec:results} and~\ref{sec:conclusion} we evaluate, compare, and discuss the proposed method using real world and virtual data.

\section{RELATED WORK}
\label{sec:related}

Verifying and identifying the limits of perception systems is crucial to ensure safe operation. Dreossi et al.~\cite{Dreossi2019}, and Volk et al.~\cite{Volk2019}, generate adversarial conditions for stress testing Convolutional Neural Networks (CNNs) in order to determine the limits of object detection.

Existing benchmarks like COCO~\cite{DBLP:journals/corr/LinMBHPRDZ14} or parts of KITTI~\cite{KITTI_benchmark} use simple performance indicators such as precision, accuracy, recall, and mean average precision (mAP) to evaluate object detection~\cite{PrecisionRecall, PrecisionRecall2, EveringhamEtAl}. 
These performance metrics are based on classifying detected objects as true positive (TP) or false positive (FP). Classification is performed on the basis of Intersection over Union (IoU) between detection and ground truth (GT).
The IoU, which is also known as Jaccard-Index, is a widespread metric to evaluate object detection~\cite{IoU}. It uses the area of the intersection and the union of a detected bounding box $D$ and the corresponding GT bounding box $G$ and is defined by Rezatofighi et al.~\cite{IoU} as

\begin{equation}
\mathrm{IoU} = \frac{|D \cap G|}{|D \cup G|}.
\end{equation}

IoU is used by object detection benchmarks like Pascal VOC~\cite{EveringhamEtAl} and COCO~\cite{DBLP:journals/corr/LinMBHPRDZ14}. The threshold value to consider an object as TP can be parameterized individually. Different threshold values like 0.5 in Pascal VOC or 0.7 in KITTI are used. They are also adjusted according to the object class to be detected.
The aforementioned metrics focus on evaluating a single frame and can be used to assess the performance of bounding box-based object detection in both 2D and 3D. However, all of these metrics consider annotated GT objects only and lack the ability to evaluate object-tracking methods.

In 2006, the \textit{Classification of Events, Activities and Relationships} (CLEAR) defined different metrics for the evaluation of object tracking, person identification, head pose estimation and acoustic-scene analysis. The results were presented by Stiefelhagen et al.~\cite{Stiefelhagen_et_al_CLEAR}. As part of the tracking evaluation, the Multiple-Object-Tracking and Multiple-Object-Detection precision (MOTP/MODP), and accuracy (MOTA/MODA) metrics were defined. Important aspects and drawbacks of these CLEAR metrics will be discussed in the following.

Let $m_t$ and $\mathit{fp}_t$ be the amount of misses and false positives at time $t$ and let $g_t$ be the number of ground truth objects at time $t$. To determine the MODP score all mapped object sets are used to calculate the IoU of each object. $N^{\mathrm{mapped}}_t$ denotes the number of mapped object sets at time $t$. Based on these values including the IoU score, MODA and MODP are defined as:

\begin{equation}
\label{MODA}
\mathrm{MODA}(t)=1-\frac{\sum_t (m_t + \mathit{fp}_t)}{\sum_t g_t},
\end{equation}

\begin{equation}
\label{MODP}
\mathrm{MODP}(t)=\frac{\sum_{i=1}^{N^{\mathrm{mapped}}_t} \mathrm{IoU}_{i}}{N^{\mathrm{mapped}}_t}.
\end{equation}

In addition to the above described values the tracking metrics include the parameter $\mathit{mme}_t$, which represents the number of mismatches between object and tracking hypothesis.
The parameter $d_{i,t}$ describes the distance between the $i^{th}$ ground truth object and tracking hypothesis at time $t$. Similar to $N^{\mathrm{mapped}}_t$, $c_t$ is the quantity of matches for a frame at time $t$. Including these additional values, MOTA and MOTP are defined as:

\begin{equation}
\label{MOTA}
\mathrm{MOTA}(t)=1-\frac{\sum_t (m_t + \mathit{fp}_t + \mathit{mme}_t)}{\sum_t g_t},
\end{equation}
\begin{equation}
\label{MOTP}
\mathrm{MOTP}(t)=\frac{\sum_{i,t} d_{i,t}}{\sum_t c_t}.
\end{equation}

Nowadays, the CLEAR metrics are an important method to evaluate the performance of object detection and tracking algorithms. They are used by relevant benchmarks such as KITTI's Multiple-Object-Tracking benchmark~\cite{KITTI_benchmark}.

The biggest advantage of the CLEAR metrics in comparison to basic performance metrics like precision and accuracy lies in the higher level of detail. Using the IoU respectively the distance to determine the precision scores allows a better statement about the precision compared to the binary way of calculation based on TP and FP amount. However, to evaluate the tracking performance only the distance to the ground truth object gets considered and the IoU is neglected. This approach might be applicable for distant objects but especially for the evaluation of closer objects the IoU has to be considered. Without an exact IoU consideration it is not possible to guarantee that autonomous vehicles correctly perceive their environment and are able to plan safe maneuvers. Hence, the CLEAR metrics, which evaluate the detection performance, have to be considered as well. However, having four single metric results does not allow for a fast evaluation.
An other drawback of the CLEAR metrics is that they only cover the evaluation of the precision and accuracy of detection and tracking without a differentiation of detection time or relevance of the object.

A new real-time performance evaluation for object detection including the CLEAR metrics has been proposed by Kim et al.~\cite{RTP16}.
As a metric for real-time video surveillance systems, they include the time which is required to detect an object. In the proposed soft real-time mode the CLEAR metric scores are mapped to an interval of $[0;\mathrm{score}]$ if detection time exceeds a given threshold $\tau$. Since the mapping function is not explicitly defined a variable usage in different systems with varying requirements in perception time is enabled. Kim et al. also proposed a hard-real-time mode in which the score is set to 0 if detection time exceeds $\tau$. 
However, in the field of video surveillance the relevance factor is less essential than in automated driving. The extension of the CLEAR metrics of Kim et al. can therefore not be directly applied to evaluate perception algorithms for autonomous vehicles. 

In 2017 Shalev-Shwartz et al.~\cite{RSS} presented a new approach to guarantee safety in automated driving. The ``Responsible-Sensitive Safety'' (RSS) model is an attempt to formalize the human judgment in different road scenarios in a mathematical sense. The RSS model consists of 34 definitions of different safety distances, times, and procedural rules to fulfill the following five simplified rules:

\begin{enumerate}
\item\label{RSS1} Do not hit anyone from behind
\item Do not cut-in recklessly
\item Right-of-way is given and cannot be enforced
\item Act carefully at areas with limited visibility
\item If it is possible to avoid a collision without causing another one, the collision must be avoided
\end{enumerate}   

These rules specify how an autonomous vehicle should behave and provide a mathematical description of a safe conduct. However, RSS does not provide an opportunity to evaluate the safety of a vehicle in a given scenario. 
If for example the environment perception of an autonomous vehicle does not detect a preceding vehicle it can not be guaranteed that rule~\ref{RSS1} of RSS will be fulfilled. To guarantee the safety of autonomous vehicles it is therefore necessary that perception systems can be evaluated whether they are able to detect all safety relevant objects in a given scenario.

Another key factor for safety of an autonomous vehicle is the ability to cope with all possible environmental influences. Different weather conditions like snow or rain affect the brake distance by decreasing the friction coefficient~\cite{BrakeaccelInfluence}. The CLEAR metrics and the metrics used for evaluation in~\cite{DBLP:journals/corr/LinMBHPRDZ14,KITTI_benchmark,EveringhamEtAl} do not consider environmental conditions. However, to evaluate the safety of autonomous vehicles these influences are of great importance and must be taken into account. Especially in adverse weather, an autonomous vehicle must be able to detect obstacles at an early stage in order to be able to react in time to longer braking distances.

\section{SAFETY METRIC}
\label{sec:metric}

The main goal of this work is to create a metric which does not only evaluate
the performance but also considers real-world safety of an object perception system.
A key requirement is to create a metric which allows
to compare different perception systems under varying road scenarios and
different weather conditions easily.
For this purpose the result has to be a single value in a defined range.

\begin{figure}[t]
  \centering
  \includegraphics[width=\columnwidth,trim=20mm 9mm 55mm 5mm,clip]{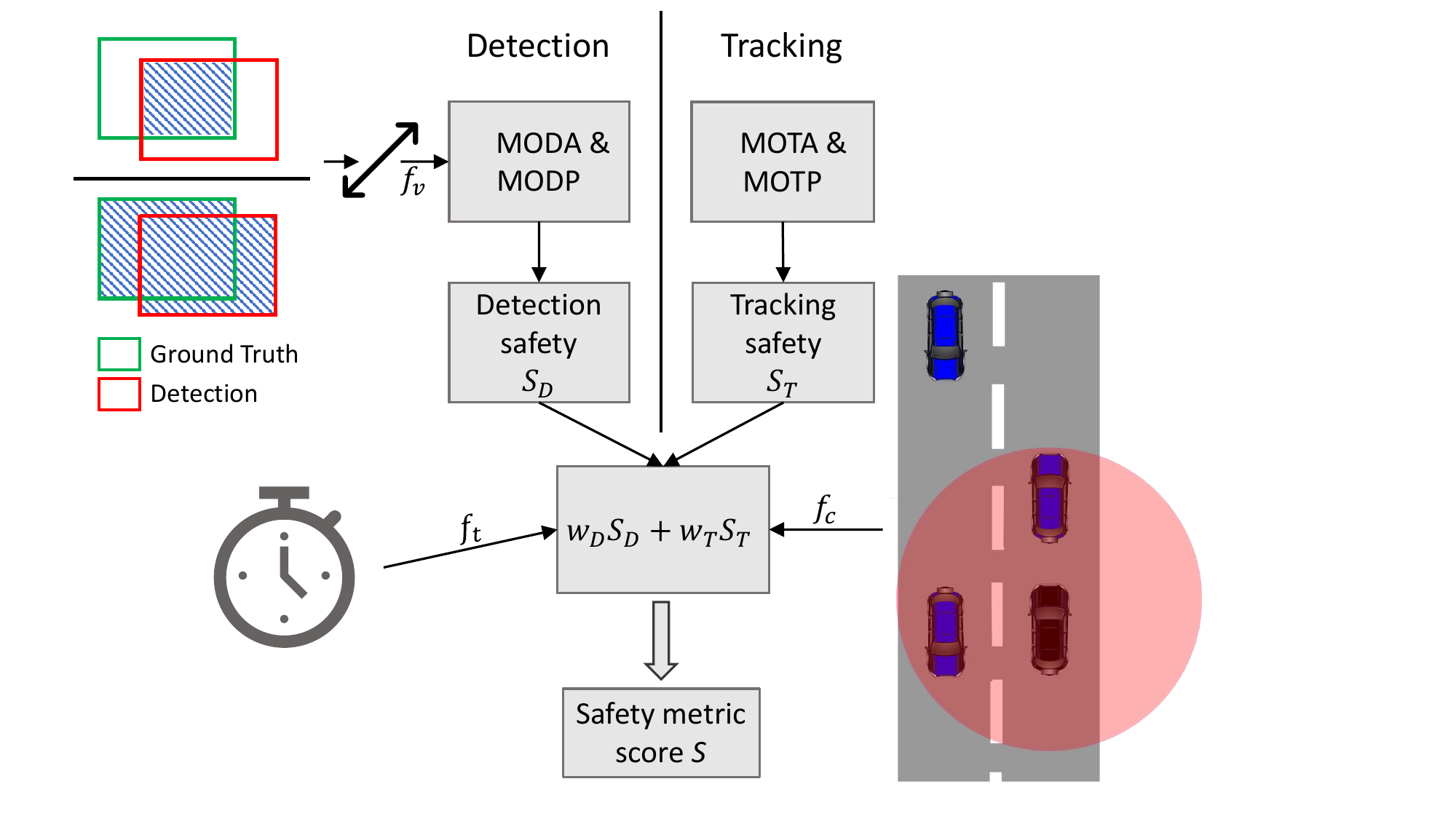}
  \caption{Process overview of the single components and their relation to one another to determine the safety metric score $S$. Ego vehicle is black. Red circle around ego indicates safety critical area.}
  \label{fig:sm-grafik}
\end{figure}

The composition of the individual safety metric components and their
relationship is presented in Fig.~\ref{fig:sm-grafik}.
It illustrates the process of how our approach combines different components to
obtain a single safety-metric evaluation that allows easy comparison of the
perceptual algorithms.

For evaluation of safety we defined three criteria to consider:

\begin{description}
\item[Quality] The quality of the perception is important for further actions to be performed such as the trajectory planning.
\item[Relevance] A potentially collision relevant object must not go
  unrecognized.
  Hence, we have to distinguish between relevant and irrelevant objects.
\item[Time] In a real-time system, time is always a relevant factor. Longer detection times lead to less reaction time and decrease the amount of possible driving maneuvers.
\end{description}

\subsection{Basis of the Safety Metric}
\label{ssec:basis}

To evaluate safety in automated driving, the quality of the object perception has an enormous relevance. Based on the perception, further actions like prediction or planning will be performed. If detection or tracking are of low quality, it could lead to erroneous planning, which could endanger the vehicle occupants and other road users.

Therefore, perception quality is a main safety factor and will be used as the basis of the safety metric. The quality consists of two factors: accuracy and precision. To combine these factors we use the CLEAR metrics~\cite{Stiefelhagen_et_al_CLEAR} as indicator for the quality of object detection and tracking (see Section~\ref{sec:related}). The choice of CLEAR metrics was based on the completeness of the metric, as it combines performance scores for multiple-object-detection and -tracking with accuracy and precision. 

Precision is defined as relation between TP and FP detections with the formula by Davis and Goadrich~\cite{PrecisionRecall2}:

\begin{equation}\label{eq:prec}
  \mathrm{Precision} = \frac{\mathit{TP}}{\mathit{TP}+\mathit{FP}}.
\end{equation}

In addition to the definition by Davis and Goadrich, MODP/MOTP indicates a more detailed precision calculation. Using Definition (\ref{eq:prec}) for precision calculation leads to a loss of information. Due to the classification as TP or FP the original overlap information of detection and ground truth is lost. MODP/MOTP preserve this information and therefore have a higher level of detail. This is another criteria for choosing the CLEAR metrics as basis of the the safety metric to evaluate the perception quality.

While using the CLEAR metrics to evaluate safety, one problem regarding the MOTP score appears. MOTP corresponds to the mean distance between tracked position and ground truth position. Hence, a lower MOTP score indicates a better tracking. This is contrary to the safety metric score where a higher score corresponds to better safety. To invert the MOTP indication, a mapping between MOTP scores and a MOTP safety metric score $\mathrm{MOTP}_s \in [0,1]$ needs to be defined. Let $T_u$ be the upper and $T_l$ the lower threshold. $\mathrm{MOTP}_s$ can be determined by using the normalization function $f_{norm}$ which is defined as
\begin{equation} \label{factor:map}
f_{norm}(x) =
\begin{cases}
  1 & x < T_l\\
  1 - \frac{x - T_l}{T_u - T_l} & T_l \leq x \leq T_u\\
  0 & \mathrm{otherwise}
\end{cases}.
\end{equation}

In our experiments we set $T_l= 0.8\,m$, as this value complies with step width of a vulnerable road users (VRU) to avoid a collision. By similar reasoning we set $T_u= 2.5\,m$, which roughly corresponds to a misjudgment that could lead to minor injuries in case of a collision.
A linear function is used because MOTP is metrically scaled. Hence, for each $\Delta \mathrm{MOTP}_1=\Delta \mathrm{MOTP}_2$, where $\Delta \mathrm{MOTP}_i$ stands for an increase of the MOTP score, the influence is equally bad, independent of the $\mathrm{MOTP}$ score. This also applies for the use of $f_{norm}$ in Section~\ref{ssec:p_time}.

In general, the threshold values of $f_{norm}$ are parameterizable based on the field of application and the corresponding demands. This increases the variability and enables the applicability of the metric for the evaluation of different systems.
For the proposed safety metric we consider precision and accuracy as equally important, thus we use the
accuracy and precision score of detection and tracking to calculate a separate
basis score. The detection safety score ($S_D$) and the tracking safety score ($S_T$) are defined as the mean of their corresponding CLEAR metrics scores:

\begin{equation}
S_D = \frac{\mathrm{MODA}+\mathrm{MODP}}{2}
,\>
S_T = \frac{\mathrm{MOTA}+\mathrm{MOTP}_s}{2}.\
\end{equation}

\subsection{Additional Criteria}
\label{ssec:criteria}
The following sections will introduce additional criteria, which are used in combination with the basis (see Section~\ref{ssec:basis}) to evaluate safety. These criteria correspond to the three safety criteria, which were introduced in Section~\ref{sec:metric}. For quality optimization we use a distance-based IoU verification. The second subsection considers the evaluation of collision relevance, which covers the criteria of relevance. Finally the detection time is considered as third criteria. 

\subsubsection{Distance-based IoU Verification}
\label{ssec:iou_ver}

During the calculation of the CLEAR metrics as basis of the safety metric a second parallel assessment is performed. For objects which are closer to the ego vehicle the perception has to be more accurate. This stricter requirement for closer objects is based on the reduced time to react in maneuver planning. Since these objects show a higher safety criticality, we have to differentiate the perception quality. The quality itself depends among others on the precision in detection (see Section~\ref{sec:metric}). This criteria is defined on basis of the IoU score.

The distance-based IoU verification uses the cover $C$ of $G$. For a detected object $o$ the cover $C_o$ is defined as
\begin{equation}
\mathit{C_o}=\frac{\mathit{{|D_o \cap G_o|}}}{\mathit{|G_o|}}\,.
\end{equation}

Using $\mathit{C_o}$, the safety function $f_s$ is defined as

\begin{equation}
f_s(C_o)= \begin{cases}
\frac{1+\mathit{mC}+(1-\mathit{mC})\sin(\pi(C_o-\frac{1}{2}))}{2} & C_o \in (\mathit{mC}, 1],\\
1 & C_o \in (1, \mathit{oT}],\\
\frac{1+\cos(\frac{\pi}{\mathit{mO}-\mathit{oT}}(C_o-\mathit{oT}))}{2} & C_o \in (oT,\mathit{mO}],\\
0 & \mathrm{otherwise}.
\end{cases}
\end{equation} 

This function guarantees a minimum detection precision $\mathit{mC}$. In between the thresholds $\mathit{mC}$ and $\mathit{mO}$, trigonometric functions are used for a smooth distance-based scaling factor depending on the precision of the detection.
$\mathit{oT}$ defines a threshold how much larger an object is allowed to be detected without lowering the detection precision. If $C_o$ is larger as $\mathit{oT}$, $\mathit{mO}$ represents the upper bound up to which $f_s$ reduces the precision towards zero. 
The second part of the verification is the mapping of $f_s(C_o)$ based on the distance of the object. To combine $f_s(C_o)$ with the corresponding distance $d_o$ between $o$ and the ego vehicle, they have to be in the same interval. Therefore the maximum distance of the evaluation range is used to normalize all distances $d_o$. This leads to $ \forall o: d_o \in [0,1]$.

The distance based score is calculated by function $g : [0,1]^2 \to [-1,1]$ where
\begin{equation}
g(x, y) = x - (1-x) \cdot (1-y).
\end{equation}

To use $g( f_s(C_o), d_o)$ as distance based detection precision factor $f_v$ it has to be transformed to be in $[0,1]$:

\begin{equation}
f_v = \frac{g( f_s(C_o), d_o ) + 1}{2}.
\end{equation}

For each detected object $o$ the IoU gets scaled by $f_v$ with $\mathrm{IoU_o} \cdot f_v$.
This additional verification of detection precision leads to a stricter rating. In context of safety, a stricter rating should be preferred.

\subsubsection{Consideration of the Collision Relevance} 
\label{ssec:coll_rel}

The second criteria for evaluating the safety of a perception system is the relevance for safety. A potentially safety critical object has a higher relevance than a non safety critical object. A distinction must therefore be made here.

An object is safety critical if its distance to the ego vehicle is less than the corresponding safety distance. To calculate the safety distance to an object we use the approach of the RSS model~\cite{RSS}. It defines different safety distances for longitudinal and lateral positioning of two objects. We use the longitudinal safety distance with same direction of movement $d_{long,s}$, with opposite direction of movement $d_{long,o}$ and the lateral safety distance $d_{lat}$~\cite[Def.\,1,\,2,\,6]{RSS}.

To evaluate the collision relevance the future position of an object needs to be predicted.
Let $v_0$ be the ego velocity and $a$ the current weather-dependent brake acceleration. The prediction time frame $t_p$ depends on the braking time $t_b$, which is defined as $t_b =\frac{v_0}{a}$. 
To ensure a sufficient prediction time, $t_p$ includes a ten percent buffer: $t_p = 1.1 \cdot t_b$.
\begin{figure*}[t]
  \centering
  \includegraphics[width=0.95\linewidth,trim=30mm 65mm 5mm 47mm,clip]{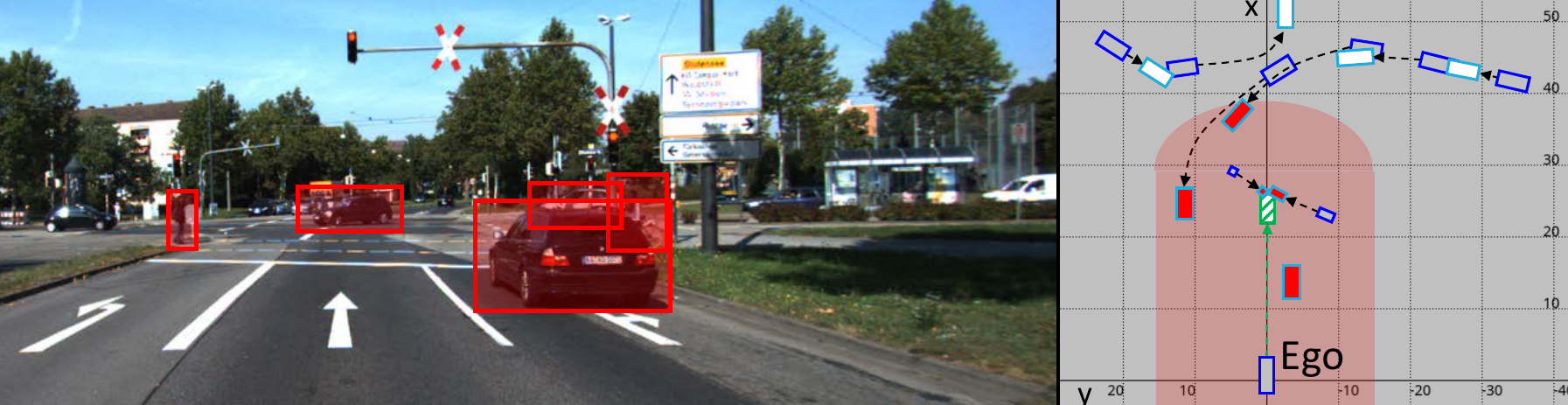}
  \caption{Schematical identification of collision relevant objects from KITTI raw dataset~\cite{KITTI_dataset}. The right image represents the bird's eye view of the camera image on the left. Blue boxes illustrate ground truth annotations, light blue boxes represent the predicted object positions from the safety metric. Red filled objects are collision relevant and white ones are not. The corresponding collision relevant objects in the camera image are marked in red.}
  \label{fig:kitti_example}
\end{figure*}

During the time span $[t_a,t_a+t_p]$, where $t_a$ represents the actual timestamp, a position prediction for each GT object and the ego vehicle is performed. For each time step it is checked whether the distance between ego and the object is greater than the corresponding RSS safety distance. If this is not the case and the object was not perceived by the perception system to be evaluated, the object is marked as safety critical.

Fig.~\ref{fig:kitti_example} shows the identification schematically. For identification of safety critical objects traffic lights as present in the illustrated scene are not considered. In order to guarantee a safe perception of the surroundings, objects that are obliged to stop at a red traffic light must also be perceived. If such a vehicle does not keep to the road traffic regulations and drives over a red light, a non-recognition would be fatal. The illustrated data represents frame 90 from drive 11 of the KITTI raw dataset~\cite{KITTI_dataset}. The velocities in this scene were increased such that a hypothetical collision would occur in the predicted time frame. The red area in the bird's eye view marks the safety critical area identified by lateral and longitudinal safety distances of RSS. The identified collision relevant objects are marked in red. If they are not detected they are safety critical as illustrated in Fig.~\ref{fig:kitti_detection}.

To rate the relevance in context of safety, we need to approximate the effect of a hypothetical collision.
The first step is an approximation of the impact velocity, in case of an in fact collision. For this purpose there are four simplified scenarios, which are head-on collision, rear-end collision, side-on collision, or a diagonal collision. Based on  this classification, the impact velocity can be approximated.

Since safety in automated driving affects not only the vehicle occupants but also other road users, these must also be taken into account. Road users can be separated into two categories. The first category are VRUs. The second category contains all road users with crush collapsible zone which includes cars, vans, trucks and trams.

The combination of impact velocity and the road user category $c$ of the
colliding object leads to a defined collision score $s_{c,ro}$ for a relevant safety critical object $ro$ which is undetected.
To determine $s_{c,ro}$ a four-level classification of the vehicle impact velocity is defined. The definition of the accident levels is based on the common accident categories used in Germany. They are defined as UK\,1 (fatality) - UK\,3 (minor injuries only) by the Ministry of the Interior of the state Nordrhein-Westfalen in Germany~\cite{Unfallkategorie}. The additional UK\,5 (material damage only) is used to include hypothetical collisions without minor injuries~\cite{Unfallkategorie}, which are nevertheless relevant.

The effects of vehicle impact velocity in a collision is investigated in works
by Frederiksson et al.~\cite{FREDRIKSSON20101672} or Han et al.~\cite{EffectsVehicleImpact}.
The defined classes with the corresponding collision scores $s_{c,ro}$ are:

\begin{equation}
  s_{c,ro} \coloneqq
  \begin{cases}
    0.9 & \textrm{no or almost no effect},\\
    0.75 & \textrm{risk of minor injuries},\\
    0.5 & \textrm{risk of serious violation},\\
    0 & \textrm{high probability of fatality}.\\
  \end{cases}
\end{equation}

For our purpose $s_{c,ro}$ is used as a factor for each frame. Therefore  $s_{c,ro} \in [0,1]$ must apply. A rated collision with high probability of fatality is unacceptable and leads to a score of 0. The case with almost no effect is worse than no accident, thus a factor of 0.9 is defined.
Due to the high effect on the final safety metric score, $s_{c,ro}$ should not be too strict otherwise the final result will drop too fast which leads to inaccuracy as there is no clear difference between the defined classes. To set $s_{c,ro}$ for the remaining two cases, the possible range of $[0,1]$ was divided by four levels which leads to $s_{c,ro}=0.75$ for minor injuries, respectively $s_{c,ro}=0.5$ for serious violations.

For each frame $t$ the minimum (worst case) $s_{c,ro}$ is calculated and used as direct relevance factor $f_c$ on $S_T$ and $S_D$. An existing collision relevance indicates a safety risk and leads to a drop of $S$.

\subsubsection{Evaluation of Perception Time}
\label{ssec:p_time}

The third criteria of a safety critical real-time perception system is the time. The more time object detection needs, the less time remains to prevent a safety critical situation. In the proposed safety metric, the time criteria gets covered by the soft real-time approach of Kim et al.~\cite{RTP16}.

For the proposed safety metric, the perception time $t_{d,o}$ of object $o$ is defined as duration from entering a safety critical area until its perception. The safety criticality is defined using the safety distances from Shalev-Shwartz et al.~\cite{RSS}, which are introduced in Section~\ref{ssec:coll_rel}. The evaluation of perception time for a perceived object with a higher distance than the corresponding RSS distance has to be adapted. For these objects we use the difference between $t_a$ and perception timestamp, which is negative. Thus an earlier perception leads to a higher safety, a positive influence of a negative detection time on the safety score is granted and valid.   

To map the detection time to a perception time factor a weighted perception time is used. The weighting was introduced because the longer the detection takes the more safety critical the situation gets. For this purpose the mean perception time is utilized to classify between long and short times. 
Let $m$ be the number of all weights and $\overline{t_d}$ the mean perception time. The weighted perception time $t_{dw}$ is then defined as
\begin{equation}
  t_{dw} =\frac{1}{m} \sum_{o}
  \begin{cases}
    t_{d,o}  & t_{d,o} \leq \overline{t_d} ,\\
    2\cdot t_{d,o}  & \mathrm{otherwise}.\\
  \end{cases}
\end{equation}

Similar as proposed by Kim et al.~\cite{RTP16}, the CLEAR metric scores are mapped by a function depending on $t_{dw}$. Mathematically equivalent $t_{dw}$ can be mapped to an interval $[0,1]$ and be used as perception time factor $f_t$.
Definition (\ref{factor:map}) with parameter $t_{dw}$ is used to determine $f_t$. $T_l$ is set to \SI{0.1}{\second}, which allows a tolerable delay in perception. For $T_u$ the ego braking time $t_b$ (see Section~\ref{ssec:coll_rel}) is used. If this time is exceeded, $f_t$ has to be 0, because an emergency stop to avoid a safety risk is not possible anymore.

\subsection{Comprehensive Safety Metric Score}
Due to the described requirement of a simple comparability between the safety
scores of different perception systems and scenarios, the safety metric is calculated as
a single score $S \in [0,1]$, where 1 is the maximum safety.
Like precision, recall, or accuracy, $S$ is calculated for a scenario with $t$ frames
using $S_D$ and $S_T$ including the evaluation of collision relevance and perception time.

To fulfill the requirement of a high variability to adapt the safety metric to various perception systems, $S_D$ and $S_T$ can be weighted variably with $w_{D}, w_{T} \in [0,1]: w_{D}+w_{T}=1$. This leads to the following safety score definition:

\begin{equation}\label{eq:SafetyScore}
  S = w_D S_D + w_T S_T.
\end{equation}

In contrast to precision, accuracy, or recall the comprehensive safety is not a percentage value, which leads to a non-intuitive interpretability. To increase the interpretability, a classification of S must be defined. The defined classification includes five levels and is based on the rating for the single CLEAR metrics values and the defined influences of collision relevance and the detection time analysis:

\begin{center}
\begin{tabular}{c    l} \toprule
    {$\textit{\textbf{S}} \boldsymbol{\in}$} & {$\textit{\textbf{Classification}}$} \\ \midrule
    $[0.0-0.2]$  & \textbf{insufficient}, high risk of fatality \\
    $(0.2-0.4]$  & \textbf{bad}, existing risk for serious violation \\
    $(0.4-0.6]$  & \textbf{good}, low probability of minor injuries  \\
    $(0.6-0.8]$ & \textbf{very good}, low risk UK\,5 collisions    \\
    $(0.8-1.0]$ & \textbf{excellent}, high probability of safe status  \\ \bottomrule
\end{tabular}
\end{center}

This classification enables a simple and quick safety evaluation of a test scenario and allows a comparison to performance metrics for more detailed analysis.

\section{RESULTS}
\label{sec:results}

To show the advantages and variability of the proposed safety metric, we compare it to widely used metrics like already discussed in Section~\ref{sec:related}.
The results are divided into three parts. In the first part we investigate an exemplary detection on the KITTI raw dataset. The second part assesses image-based object detection performance of the three well-known neural networks RRC~\cite{Ren2017}, Faster-RCNN~\cite{Ren2016}, and YOLOv3~\cite{yolov3}. In the last part the metric is used to evaluate a 3D object detection and tracking system. An IoU threshold of 50\% was used to classify an object as TP detection for both image-based and 3D object detection.

For evaluation real and virtual data is used. The evaluation was conducted without environmental influences like rain or snow. However, first evaluations have shown that the final safety score decreases due to consideration of reduced road friction.
For real world data KITTI's raw data recordings~\cite{KITTI_dataset} were utilized. The scenes provide labeled 3D object tracklets and odometry information for the ego vehicle. This data is required for the safety metric. Otherwise, the safety specific evaluation as shown in Section~\ref{sec:metric}-B can not be conducted.
Only those recordings of the KITTI raw dataset were used for evaluation which had a length of at least 30 seconds and at least 10 labeled objects. In different KITTI benchmark tasks like 3D or 2D object detection each object label gets evaluated separately. For the evaluation of the safety metric and comparison with other performance metrics all present labels in the recorded data were considered and evaluated. 

As source for virtual test data Vires VTD~\cite{vtd} was used. Three different scenarios as presented by Volk et al.~\cite{volk_environment-aware_2019} were considered. An urban crossing scenario with a lot of obstructions, a motorway scenario with higher velocities, and a rural one containing only few road users on a country road.

\subsection{Exemplary KITTI scene}

\begin{figure}[t]
  \centering
  \includegraphics[page=2,width=0.6\linewidth,trim=237mm 60mm 5mm 43mm,clip]{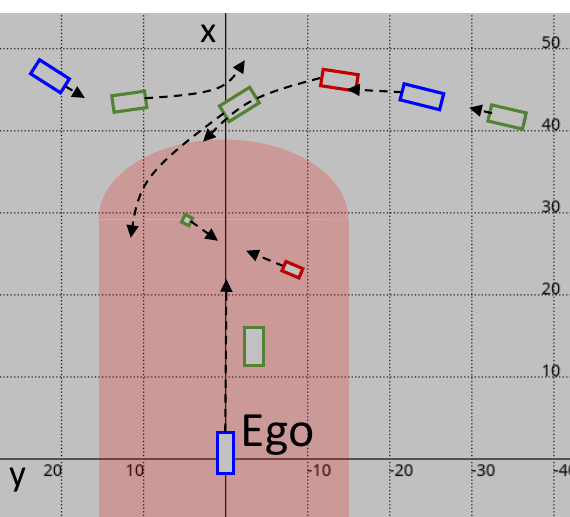}
  \caption{Exemplary bird's eye view detection on KITTI raw dataset scene as in Fig~\ref{fig:kitti_example}. Blue boxes represent ground truth, green ones are correctly detected objects and red ones are safety critical objects.}
  \label{fig:kitti_detection}
\end{figure}

An exemplary KITTI scene gets evaluated at first to show the behaviour of the proposed safety metric. The scene is illustrated in Fig.~\ref{fig:kitti_detection} and is the same as in Fig.~\ref{fig:kitti_example} which additionally shows the corresponding camera image. Fig.~\ref{fig:kitti_detection} shows TP detections in green and missed detections which where identified as safety critical in red. Blue boxes mark remaining ground truth objects which are not yet safety critical. This scene contains five TP detections, two missed safety critical detections and two missed detections excluding the Ego vehicle. This results in a recall value $rec=\frac{5}{9}=55.\overline{5}\%$ and a precision $prec=\frac{5}{5}=100\%$. For the collision relevance evaluation (see Section~\ref{ssec:coll_rel}) the not detected cyclist is a VRU. Including the velocity this leads to a relevance factor $f_c=0$ as a collision with a VRU has a high probability of fatality. The final safety metric score $S$ would therefore be 0. This shows that existing metrics do not capture the severity of missed VRU detections but the proposed safety metric does.

\subsection{Image-based Object Detection}

For image-based object detection three neural networks are evaluated. The results of the proposed safety metric are compared against the currently used evaluation metrics recall, precision, and mean average precision as presented in Section~\ref{sec:related}. The results of the different metrics for YOLOv3, Faster-RCNN, and RRC are shown in Tables~\ref{tab:res_image_yolo}, \ref{tab:res_image_faster}, and~\ref{tab:res_image_rrc}. The results for KITTI should not be mistaken with the online available results on the KITTI benchmark page. We performed the evaluation on all class labels while they are listed separately per class label on the KITTI benchmark site. Therefore, the presented results show lower values.

The final safety score $S$ (see Eq.~\ref{eq:SafetyScore}) is composed of $S_D$ and $S_T$. Image-based object detection alone does not allow for an evaluation of tracking performance. Therefore, $w_T$ was set to 0. This way it is possible to adapt the safety metric to the specific target system under evaluation. With $w_T=0$ the final safety score $S=S_D$.

When we compare the results for the three different algorithms, it can be seen that the safety score has a similar trend as mAP. However, in some examples, like Faster-RCNN on the KITTI benchmark, the safety score is much lower than the mAP. This indicates that Faster-RCNN was not able to detect collision relevant objects precisely or completely missed to detect them. In the KITTI case RRC performed best according to the standard performance metrics. This got confirmed with the safety metric which made the advantage of RRC over Faster-RCNN even more clear.

The results on the virtual scenarios allow a more focused evaluation as there is a complete ground truth. The safety score on the motorway scenario is higher than mAP in all cases except for YOLOv3.
This is due to the fact that the safety metric considers the higher velocities of the scene and therefore harshly penalizes missed detections. This indicates that YOLOv3 on the one hand detects more objects that are less relevant or precise. On the other hand, RRC detected the least number of objects in the motorway scenario but it also detected the most collision relevant ones.

For the rural scenario, all algorithms show a much better performance according to the safety score than the other performance metrics because this scenario contains only two objects. A far distant object, which hardly got detected by any algorithm and is not relevant for safety. This shows that the safety metric automatically adapts to a specific scene and automatically gives an indication how well an algorithm performed in perceiving the current environment.

\begin{table}[t]
\centering
\caption{Evaluation results for object detection with YOLOv3.}
\begin{tabular}{l SSSS }

\toprule
& & \multicolumn{3}{c}{Virtual Scenarios}   \\
\cmidrule{3-5}
& KITTI &  Motorway & Crossing & Rural \\
\midrule

Precision              &  0.59 &  0.82 &   0.86 &  0.96\\
Recall                 &  0.60  &  0.23 &   0.36 & 0.60 \\ 
mAP                    &  0.51  & 0.21  &  0.35  & 0.60 \\
\midrule
\addlinespace
Safety score $S$       &  0.48  & 0.14  &  0.20  & 0.78 \\
\addlinespace
\bottomrule
\end{tabular}
\label{tab:res_image_yolo}
\end{table}

\begin{table}[t]
\centering
\caption{Evaluation results for object detection with Faster-RCNN.}
\begin{tabular}{l SSSS }

\toprule
& & \multicolumn{3}{c}{Virtual Scenarios}   \\
\cmidrule{3-5}
& KITTI &  Motorway & Crossing & Rural \\
\midrule

Precision              &  0.58  & 0.57  &  0.43  &  0.59\\ 
Recall                 &  0.73  &  0.10 &  0.19  & 0.36 \\ 
mAP                    &  0.64  &  0.09 &  0.12  &  0.30\\
\midrule
\addlinespace
Safety score $S$       &  0.46  & 0.11  &  0.14  &  0.52\\
\addlinespace
\bottomrule
\end{tabular}
\label{tab:res_image_faster}
\end{table}

\begin{table}[t]
\centering
\caption{Evaluation results for object detection with RRC.}
\begin{tabular}{l SSSS }
\toprule
& & \multicolumn{3}{c}{Virtual Scenarios}   \\
\cmidrule{3-5}
& KITTI &  Motorway & Crossing & Rural \\
\midrule
Precision              &  0.86  & 0.89  &  0.64  & 1.00 \\ 
Recall                 &  0.73  & 0.04  &  0.12  &  0.14\\ 
mAP                    &  0.69  &  0.04 &  0.09  &  0.14\\
\midrule
\addlinespace
Safety score $S$       &  0.69  &  0.19 &  0.16  &  0.79\\
\addlinespace
\bottomrule
\end{tabular}
\label{tab:res_image_rrc}
\end{table}

\subsection{3D Object Detection and Tracking}

For evaluating 3D object detection and tracking the perception pipeline presented by Volk et al.~\cite{volk_environment-aware_2019} was employed. The pipeline was extended to use YOLOv3 as image-based object detection in combination with the L-shape fitting algorithm by Zhang et al.~\cite{zhang_efficient_2017}. 
For the KITTI dataset the lidar-based object detection PointPillars~\cite{lang_pointpillars_2019} integrated by Shi et al.~\cite{shi2020points} is utilized. The 3D object detection is followed by a Kalman Filter using a constant velocity model. 

It is beyond the scope of this paper to provide an optimal 3D detection and tracking, but it does show the impact of different environments on the proposed safety metric. Furthermore, only vehicle-local perception will be evaluated.
Similar to the image-based evaluation, the results for KITTI should not be mistaken with the online available results as we evaluated all class labels.

For the evaluation we set detection and tracking as equally important with $w_{D}=0.5$ and $w_{T}=0.5$. The results of the evaluation are shown in Table~\ref{tab:res_3d_yolo}.
The metrics evaluated using the KITTI raw data show similar trends. The safety score $S$ is higher than the metric values mAP and MODA. This is a consequence of the safety metric focusing on critical objects which might lead to a collision if not detected. The original CLEAR metric for tracking accuracy MOTA is higher than $S_T$. The tracking precision MOTP is almost \SI{5}{\meter} worse compared to the virtual scenarios as all KITTI object classes are evaluated. Especially trams represent very long objects resulting in a fast decrease of tracking precision if not correctly detected. This lowers the tracking safety $S_T$. However, tracking safety $S_T$ is still higher compared to detection safety $S_D$. 

For the virtual motorway scenario higher velocities and larger breaking distances show the lowest safety score $S$. $S$ is still higher as for example recall, mAP, or MODA. However, the difference between mAP and $S$ is the smallest of all investigated virtual scenarios including the KITTI dataset. This is because the highway scenario contains more collision relevant objects dense traffic and distant objects, which are mostly occluded by other traffic participants.

The intersection scenario behaves similarly to the motorway scenario, with the difference that occlusions are not caused by other objects but by the environment. In this case the safety score is higher compared to mAP and MODA because objects exiting the crossing are less relevant as a possible collision is unlikely compared to entering traffic.

The safety score for the rural road scenario shows the highest results in the virtual scenes compared to detection evaluation with precision, recall, mAP, MODA, or MODP. Again the crash relevant objects are in focus by the safety metric. The hard to detect more distant object is not considered because the safety distances according to the definitions by RSS are large enough.

\begin{table}[t]
\centering
\caption{Bird's eye evaluation results for 3D object detection and tracking.}
\begin{tabular}{l SSSS }
\toprule
& & \multicolumn{3}{c}{Virtual Scenarios}   \\
\cmidrule{3-5}
& KITTI &  Motorway & Crossing & Rural \\
\midrule
Precision              &  0.50  &  0.32  &  0.43  &  0.52\\
Recall                 &  0.50  &  0.13  &  0.18  &  0.35 \\ 
mAP                    &  0.37  &  0.09  &  0.12  &  0.22\\
\midrule
MODA                   &  0.35  &  0.05 &  0.08  &  0.13\\  
MODP                   &  0.48  &  0.39 &  0.48  &  0.44 \\  
MOTA                   &  0.65  &  0.39 &  0.28  &  0.72\\  
MOTP [m]               &  6.83  &  1.64 &  1.36  &  1.87 \\
\midrule
$S_D$                  &  0.42  &  0.07 &  0.14  &  0.32\\ 
$S_T$                  &  0.53  &  0.19 &  0.24  &  0.83 \\
\midrule
\addlinespace
Safety score $S$       &  0.47  &  0.13 &  0.19  &  0.58\\
\addlinespace
\bottomrule
\end{tabular}
\label{tab:res_3d_yolo}
\end{table}


\section{CONCLUSION \& OUTLOOK}
\label{sec:conclusion}

In this paper we presented a new way to assess safety of object perception systems. In contrast to currently used evaluation metrics like mean average precision or the CLEAR metrics the presented metric allows for a more complete evaluation.
The proposed metric identifies collision relevant zones with definitions by RSS and penalizes undetected collision relevant objects. Real time aspects are incorporated by considering detection times as well as an additional detection quality check. 
A key advantage of the proposed metric comes due its variability. The metric can be used for evaluating image-based 2D object detection, 3D object detection, single and multiple-object-tracking systems, and even cooperative perception systems.

The safety metric always considers one specific ego vehicle as evaluation target for which the safety score gets evaluated. This score does not guarantee the safety of an autonomous systems as this metric is only intended to evaluate object perception. However, prediction and planning are equally important. The safety value is intended to evaluate safety under the assumption that you must be aware of all potentially dangerous objects in your environment in order to initiate collision avoidance maneuvers.
The result is always one single and easily comparable value which is categorized into five simplified safety stages (insufficient, bad, good, very good and excellent). This simplifies the interpretation and allows for a fast safety assessment.

As a next step, we are going to further improve our metric and use it to evaluate the safety of vehicle-local and cooperative perception under adverse environmental conditions.


\section*{ACKNOWLEDGMENT}

This work has been partially funded by the German Research Foundation (DFG) in the priority program 1835 under grant BR2321/5-2.



\bibliographystyle{IEEEtran} 
\bibliography{literature}

\end{document}